\documentclass[11pt]{article}

\usepackage[preprint]{acl}
\usepackage{times}
\usepackage{latexsym}
\usepackage[T1]{fontenc}
\usepackage[utf8]{inputenc}
\usepackage{microtype}
\usepackage{inconsolata}
\usepackage{array}
\usepackage{booktabs}
\usepackage{graphicx}
\usepackage{longtable}

\newcolumntype{L}[1]{>{\raggedright\arraybackslash}p{#1}}

\title{A Komi-Yazva--Russian Parallel Corpus and Evaluation Protocol for Zero- and Few-Shot LLM Translation}

\author{Petr Parshakov \\
  HSE University, Perm, Russia \\
  School of Management SKOLKOVO, Moscow, Russia \\
  \texttt{parshakov.petr@gmail.com}}

\begin{document}
\maketitle

\begin{abstract}
We present the first Komi-Yazva--Russian parallel corpus together with an explicit evaluation protocol for studying LLM translation in an endangered, extremely low-resource setting. The dataset contains 457 aligned sentence pairs from 74 narrative texts and is accompanied by documented provenance, sentence-level alignment, and story identifiers that enable leakage-aware evaluation. Our contribution is therefore not only a new dataset, but a reproducible evaluative setup: story-level cross-validation, deterministic retrieval for few-shot prompting, strict validation of generated outputs, complementary reference-based and judge-based metrics, and story-level uncertainty estimates. We use this setup to compare modern large language models on Komi-Yazva-to-Russian translation under severe parallel-data scarcity in zero-shot and retrieval-based few-shot regimes. Across models, LLMs produce non-trivial translations, but performance varies strongly by model family and by prompting regime. Retrieval-based few-shot prompting consistently improves over zero-shot prompting, while gains beyond a small retrieved context remain limited, suggesting that most of the accessible benefit comes from leaving the zero-shot regime rather than from continually increasing the number of demonstrations. At the same time, quality-oriented metrics, edit-based metrics, judge-based scores, and failure-aware reliability counts do not always privilege exactly the same systems, which shows that evaluative conclusions in this setting depend materially on what is measured and how failures are handled. We therefore frame the paper as both a dataset contribution and an evaluation study: the corpus, protocol, and explicit statement of scope and limitations together provide a foundation for more reproducible and interpretable assessment of endangered-language translation.
\end{abstract}

\section{Introduction}
Machine translation for extremely low-resource languages remains difficult because sparse supervision interacts with limited lexical coverage, domain mismatch, and unstable prompting behavior. These problems are especially acute for endangered and underrepresented language varieties, where even basic parallel data may be missing. In the Uralic context, existing computational resources are concentrated on Komi-Zyrian and Komi-Permyak, while Komi-Yazva remains largely absent from translation benchmarks and reproducible MT evaluation \citep{rueter-etal-2020-komi-permyak,partanen-etal-2018-ud-komi,tereshchenko-etal-2025-gpt-uralic}.

This paper addresses that gap as a combined dataset and evaluation contribution. First, we introduce, to our knowledge, the first parallel Komi-Yazva--Russian corpus and document the scope, assumptions, and intended evaluative use of the data. Second, we pair the corpus with a controlled evaluation protocol for extremely low-resource translation. The protocol is designed to support reproducible claims about relative model behavior under story-level generalization, rather than broad claims about open-domain translation quality. Concretely, we compare zero-shot prompting with retrieval-based few-shot prompting in order to test whether general-purpose LLMs can provide meaningful translation support from Komi-Yazva into Russian when only a very small amount of parallel data is available.

Our results show that LLMs can produce non-trivial Russian translations from Komi-Yazva, but that the conclusions depend on the evaluation lens. Retrieval-based few-shot prompting consistently improves over zero-shot prompting, whereas increasing the number of demonstrations beyond a small few-shot context yields only limited additional gains. At the same time, quality-oriented metrics, failure-aware reliability counts, and judge-based assessments do not always privilege exactly the same systems. Together, these findings position the new corpus not only as a resource contribution, but also as a case study in how evaluation design shapes scientific claims in an endangered-language setting.

Accordingly, the paper should be read not as a new-model paper, but as an evaluation-and-dataset paper. Its main contributions are: (i) a new parallel Komi-Yazva--Russian dataset with documented provenance and explicit evaluative scope; (ii) a reproducible evaluation protocol with story-level leakage control, deterministic retrieval, strict generation validation, and story-level uncertainty estimation; and (iii) an empirical comparison of zero- and few-shot LLM translation that reveals both quality gains from retrieval and meaningful trade-offs across evaluation criteria.

The remainder of the paper reviews the relevant literature, introduces the corpus, defines the evaluation protocol, reports the evaluation results, and then discusses their implications and limitations.

\section{Literature Review}
\label{sec:lrl-llm-mt}

Research on extremely low-resource machine translation has shifted from purely supervised neural baselines toward prompting-centric use of general-purpose large language models. In high-resource directions, prompting can already be competitive, but low-resource and non-English-centric directions still show large variance and frequent failure modes \citep{hendy-etal-2023-gpt}. This motivates a more careful view of \emph{which} prompting designs work in truly data-scarce settings and when fine-tuning or multilingual transfer remains preferable.

\subsection{Komi and Komi-Yazva context}
Within Uralic NLP, Komi is represented primarily by work on Komi-Zyrian and Komi-Permyak infrastructure rather than direct Komi-Yazva MT evaluation. Available studies report foundational resources: spoken/written Komi corpus annotation workflows in ELAN \citep{gerstenberger-etal-2017-elan-komi}, Universal Dependencies treebanks for Komi-Zyrian \citep{partanen-etal-2018-ud-komi}, computational infrastructure adaptation for Komi-Permyak \citep{rueter-etal-2020-komi-permyak}, and speech-recognition baselines for low-resource Komi \citep{hjortnaes-etal-2020-speech-komi}. Together, these works show that the ecosystem for endangered Komi varieties is active, but mostly oriented toward documentation, morphology, parsing, and ASR.

Translation-specific LLM evidence for Komi is still limited but emerging. \citet{tereshchenko-etal-2025-gpt-uralic} evaluate OpenAI GPT models for endangered Uralic translation directions that include Komi-Zyrian, comparing reasoning and non-reasoning model families and analyzing refusal behavior. Their findings indicate that model architecture and prompting style materially affect whether systems attempt translation at all, which is highly relevant for endangered-language deployment.

At the same time, a dedicated benchmark for translation from Komi-Yazva into Russian with controlled zero- and few-shot LLM evaluation remains largely absent in the literature covered here. This creates a clear gap for corpus construction and targeted evaluation: a new parallel Komi-Yazva--Russian corpus can support reproducible experiments and complement existing Komi-Zyrian/Permyak-centric resources.

\subsection{Prompting in extremely low-resource and no-resource settings}
Initial evidence from underrepresented and indigenous languages suggests that LLMs can deliver non-trivial translation quality even when direct bitext is minimal, especially when prompts encode explicit lexical constraints \citep{elsner-needle-2023-translating,liao-etal-2024-learning}. In no-resource regimes, where available supervision can fall below 100 sentence pairs, comparative studies report that in-context prompting can outperform straightforward low-resource fine-tuning pipelines that rely on insufficient parallel corpora \citep{thakur-2024-no-resource}. This is an important distinction: for the \emph{extreme} tail, inference-time adaptation may be more practical than parameter updates.

Recent work on unseen-language adaptation further reinforces this point. Frameworks that combine dictionary cues with structured in-context examples can bootstrap translation into languages poorly represented or absent in pretraining, often from only a few thousand aligned examples \citep{zhang-etal-2024-unseen}. These findings support the idea that explicit task framing at inference time can compensate for missing language coverage to a meaningful degree.

\subsection{Retrieval and example selection as core design choices}
Across studies, one robust trend is that prompt quality depends heavily on demonstration selection. Retrieval-aware few-shot prompting consistently outperforms random-example prompting in low-resource settings, with gains tied to semantic similarity and better lexical match \citep{zebaze-etal-2025-context}. Compositional prompting and decomposition-based prompting provide additional improvements when source inputs are long or structurally complex, and may improve out-of-domain behavior \citep{zebaze-etal-2025-compositional}. Case studies on Mambai and other underrepresented languages similarly show that dictionary entries plus retrieved examples are often the strongest practical recipe under tight data constraints \citep{merx-etal-2024-mambai}.

\subsection{Prompting versus fine-tuning and transfer baselines}
Despite strong few-shot gains, empirical comparisons usually show that when even modest high-quality bitext is available, supervised adaptation remains very competitive or superior. Studies comparing prompting, parameter-efficient tuning, and full fine-tuning find that prompting alone rarely dominates across directions \citep{alves-etal-2023-steering,zhang-etal-2023-machine}. However, prompting remains valuable as a data amplifier: dictionary-guided prompting and synthetic example generation can improve downstream fine-tuning performance in low-resource regimes \citep{pei-etal-2025-understanding,ghazvininejad-etal-2023-dictionary}.

Transfer-based baselines are therefore still essential in any serious comparison. Multilingual pretraining and cross-lingual transfer continue to provide strong performance anchors for under-resourced languages \citep{kumar-etal-2021-machine,nllb-team-2022-no}. For language pairs with little or no direct supervision, pivot-based methods and synthetic pivoting remain practical and often strong baselines \citep{ahmed-buys-2024-neural}. Prompt-based pivoting with related languages can help, but reported gains are often sensitive to pivot choice and demonstration construction \citep{ramasethu-etal-2026-pivot}.

\subsection{Empirical studies aligned with underrepresented-language translation}
The literature spans several complementary lines of evidence. Work on endangered Uralic translation highlights the importance of model family and prompting behavior for practical usability \citep{tereshchenko-etal-2025-gpt-uralic}, while Komi-focused resource papers show that infrastructure is stronger for Zyrian and Permyak than for Komi-Yazva itself \citep{rueter-etal-2020-komi-permyak,partanen-etal-2018-ud-komi}. In no-resource and very low-resource settings, in-context prompting, dictionary grounding, and retrieval-augmented prompting often outperform naive zero-shot setups and can even be preferable to direct low-data fine-tuning \citep{thakur-2024-no-resource,merx-etal-2024-mambai,elsner-needle-2023-translating,zhang-etal-2024-unseen,zebaze-etal-2025-context,zebaze-etal-2025-compositional}. At the same time, once bitext becomes somewhat less scarce, supervised adaptation, multilingual transfer, and pivot-based methods remain strong comparison points \citep{alves-etal-2023-steering,zhang-etal-2023-machine,ahmed-buys-2024-neural,ramasethu-etal-2026-pivot}. For completeness, a structured summary of representative studies is moved to Appendix~\ref{app:lrl-matrix}.

Overall, the literature converges on a layered picture. First, in truly extreme settings (few hundred pairs or less), zero-shot prompting is usually unreliable, while retrieval-aware and dictionary-grounded few-shot prompting can provide meaningful translation quality \citep{thakur-2024-no-resource,merx-etal-2024-mambai,zebaze-etal-2025-context}. Second, once data grows toward the low-thousands, fine-tuning and multilingual transfer often regain an advantage, especially when combined with synthetic data or lexical grounding derived from prompting \citep{pei-etal-2025-understanding,alves-etal-2023-steering,zhang-etal-2023-machine}. Third, for non-English transfer and pivot scenarios, prompt-only strategies should be evaluated against strong multilingual and pivot baselines, because these methods remain highly competitive in low-resource conditions \citep{nllb-team-2022-no,ahmed-buys-2024-neural,ramasethu-etal-2026-pivot}. Finally, recent work indicates that reasoning-heavy prompting and adaptation strategy selection can further shift outcomes, suggesting that low-resource suitability depends not only on model scale but on careful inference-time design choices \citep{frontull-stroehle-2025-reasoning,toukmaji-flanigan-2025-adapting}.
For Komi specifically, current evidence is uneven: resource-building is comparatively strong for Zyrian/Permyak, while MT evaluation is still sparse and rarely centered on Komi-Yazva \citep{rueter-etal-2020-komi-permyak,partanen-etal-2018-ud-komi,gerstenberger-etal-2017-elan-komi,tereshchenko-etal-2025-gpt-uralic}. Therefore, constructing a Komi-Yazva--Russian parallel corpus together with an explicit evaluation protocol for zero- and few-shot LLM translation is a timely and well-motivated contribution that directly addresses both a resource gap and an evaluation gap. The next section describes that corpus.

\section{Corpus Description}
\label{sec:corpus}

To address the gap identified in Section~\ref{sec:lrl-llm-mt}, we construct a parallel Komi-Yazva--Russian corpus consisting of 457 aligned sentence pairs extracted from 74 narrative texts (stories). Each instance corresponds to a sentence-level alignment between a source sentence in Komi-Yazva and its Russian translation.

The corpus is constructed from the linguistic monograph \textit{Komi-Yazvinsky Dialect} by V.~I.~Lytkin \citep{lytkin1961komi}, which contains narrative texts and examples collected during fieldwork among Komi-Yazva speakers in the Perm region. The original material was digitized and manually processed to obtain sentence-level alignments between Komi-Yazva and Russian, resulting in a parallel corpus suitable for computational analysis.

The corpus is organized using two identifiers: \texttt{id\_story}, which denotes the narrative text, and \texttt{id\_paragraph}, which indexes sentences within each story. This organization supports grouping at the document level and enables story-level train/dev/test splitting in downstream experiments, reducing the risk of leakage across splits.

\subsection{Corpus Statistics}

Table~\ref{tab:corpus_stats} summarizes the main dataset characteristics. The corpus contains 457 sentence pairs distributed across 74 stories, with an average of 6.18 sentences per story. The distribution is highly skewed: some stories contain a single sentence, whereas the largest story contains 75 sentences.

Sentence lengths are short. Under whitespace tokenization, the mean sentence length is 5.55 tokens for Komi-Yazva and 5.92 tokens for Russian.

\begin{table}[t]
\centering
\small
\begin{tabular}{l c}
\hline
\textbf{Statistic} & \textbf{Value} \\
\hline
Sentence pairs & 457 \\
Stories & 74 \\
Avg. sentences per story & 6.18 \\
Min / max sentences per story & 1 / 75 \\
Avg. length (Komi-Yazva) & 5.55 tokens \\
Avg. length (Russian) & 5.92 tokens \\
\hline
\end{tabular}
\caption{Summary statistics of the Komi-Yazva--Russian parallel corpus.}
\label{tab:corpus_stats}
\end{table}

\subsection{Data Characteristics}

The dataset is an \textit{extremely low-resource} parallel corpus, with fewer than 500 aligned sentence pairs. Such settings are common for minority and endangered languages, where large-scale digital resources are typically unavailable.

The corpus has several notable properties. First, sentences are relatively short, which is consistent with narrative and conversational discourse and lower sentence-level syntactic complexity. Second, the document-level structure (stories) supports evaluation beyond isolated sentences and enables more realistic experimental protocols. Third, alignment is strictly one-to-one, with each Komi-Yazva sentence paired with a single Russian translation.

The corpus reflects translation between Komi-Yazva, a severely under-resourced Uralic language with rich morphology, and Russian, a well-resourced Indo-European language. This typological and resource asymmetry introduces several MT challenges, including morphological divergence, lexical gaps, and possible differences in preferred word order and grammatical realization.

Given its extremely small size, the corpus is not suitable for training conventional neural MT systems from scratch. Instead, it serves as a focused testbed for alternative strategies, especially zero-shot and few-shot prompting with large language models.

More broadly, the dataset enables targeted analysis of model behavior and evaluation metrics under severe data scarcity---a setting still underexplored in current MT literature. As such, the corpus provides a rare opportunity to study translation in a microlanguage scenario where both linguistic resources and digital representation are highly limited, which is relevant for inclusive and equitable NLP. Just as importantly, the dataset is accompanied by an explicit statement of what it is meant to evaluate: relative model quality, robustness, and prompt sensitivity on unseen stories within this narrative-domain corpus. Section~\ref{sec:experimental_design} turns this corpus into a controlled evaluation protocol for zero-shot and retrieval-based few-shot LLM translation from Komi-Yazva into Russian.

\subsection{Intended Evaluative Use and Claims}

This dataset is intended primarily as an evaluation resource rather than as a training corpus for large-scale model development. In its current form, it is designed to support relative claims about how prompting regimes and model families behave under extreme parallel-data scarcity when tested on unseen stories. The setup assumes sentence-level translation, one-to-one sentence alignments, narrative-domain source texts, and a single Russian reference per Komi-Yazva sentence.

These assumptions place important bounds on interpretation. The dataset can meaningfully support claims about comparative translation quality, operational reliability under strict output validation, and sensitivity to retrieved demonstrations within this task formulation. It does not by itself justify claims about open-domain Komi-Yazva translation, discourse-level translation quality, human deployment readiness, or the overall multilingual competence of the evaluated models. Making these intended uses explicit is part of the contribution, because in a dataset this small, evaluation design and interpretive scope are inseparable.

To support direct reuse and verification, the released dataset is available on \href{https://huggingface.co/datasets/pparshakov/komi-yazva-russian-parallel}{Hugging Face}, and the corresponding code and evaluation pipeline are available on \href{https://github.com/pparshakov/komi-yazva}{GitHub}.

\section{Evaluation Protocol}
\label{sec:experimental_design}

We treat evaluation methodology itself as part of the contribution. We study translation from Komi-Yazva into Russian in an extremely low-resource setting, where the corpus is small, the source language is morphologically rich, and test examples from the same narrative can share lexical material, named entities, and local stylistic regularities. For this reason, our protocol emphasizes four properties: \emph{leakage control}, \emph{robustness to split variation}, \emph{comparability across prompting regimes and model families}, and \emph{clarity about which claims the setup can support}.

\paragraph{Evaluative claims supported by the protocol.}
The protocol is designed to support relative claims about (i) quality differences across model families, (ii) the effect of retrieval-based few-shot prompting relative to zero-shot prompting, and (iii) quality--reliability trade-offs once malformed generations are counted explicitly. It is not designed to certify absolute translation adequacy for deployment, nor to determine a universally best multilingual model outside this dataset. This distinction matters because, in a low-resource setting, conclusions can change materially with split design, metric choice, and failure accounting.

\paragraph{Sentence-level translation with story-aware evaluation.}
The basic prediction unit in our experiments is a single sentence. Each Komi-Yazva sentence is translated independently into Russian, and models are not given explicit discourse context from neighboring sentences at inference time. We adopt sentence-level prediction for two reasons. First, it keeps prompts short and stable across models with different context-management behavior. Second, it allows us to isolate the effect of in-context examples without conflating it with longer-context narrative modeling. At the same time, because the corpus is organized into stories, we preserve story structure in the split protocol and in the uncertainty estimates.

\paragraph{Story-level folds.}
A sentence-level random split would be inappropriate in this setting. Stories in the corpus frequently contain repeated lexical material, recurring referents, and highly similar local constructions. If some sentences from a story appear in training and other sentences from the same story appear in test, the evaluation becomes overly optimistic: models may effectively benefit from narrative overlap rather than true generalization to unseen stories. To avoid this, we use $5$-fold cross-validation with \texttt{GroupKFold}, grouping by story identifier. All sentences from a story are assigned to the same fold, so the test set in each fold contains only stories that were entirely unseen when selecting few-shot examples.

We use multiple folds rather than a single train/test split because the corpus is small enough that results can be sensitive to the particular subset of stories held out. Evaluating across five story-level folds reduces variance due to a single arbitrary partition and yields a more reliable estimate of model behavior across different subsets of the narrative material. In this sense, the folds are not only a way to maximize data usage, but also a way to measure robustness under changing held-out story composition.

\paragraph{Multiple values of $k$.}
For each test sentence, we evaluate three prompting regimes: zero-shot ($k=0$) and two few-shot conditions ($k=4$ and $k=8$). This design is meant to answer a specific question: in a genuinely low-resource scenario, how much can in-context learning compensate for the lack of task-specific supervised training?

The zero-shot condition establishes a lower-bound baseline for raw multilingual transfer and general instruction following. The two few-shot conditions then probe whether the model can exploit a very small number of aligned examples from the training split. We do not treat the choice of $k$ as arbitrary. A very small $k$ is necessary because the corpus itself is small and because examples must remain relevant rather than simply filling the context window. We therefore choose $k=4$ as a minimal but nontrivial few-shot condition, and $k=8$ as a stronger in-context condition that still fits comfortably within prompt budgets for all tested models. Comparing $k=0$, $k=4$, and $k=8$ lets us examine whether gains from demonstrations are monotonic, whether they saturate quickly, and whether some models are more sensitive than others to the number of in-context exemplars.

\paragraph{Deterministic retrieval of few-shot examples.}
Few-shot examples are never sampled randomly. Instead, for each fold we construct a retrieval index over the source side of the training partition using TF--IDF over character-window $n$-grams (\texttt{analyzer=char\_wb}, \texttt{ngram\_range=(3,5)}). For each test sentence, we retrieve the top-$k$ most similar source sentences from the training split and insert their aligned Russian references into the prompt.

We prefer this deterministic retrieval strategy over random sampling for both methodological and linguistic reasons. Methodologically, it makes the prompting procedure fully reproducible: for a given fold, model, and test sentence, the few-shot context is fixed. Linguistically, character-$n$-gram matching is better suited than purely word-level matching to a morphologically rich, low-resource language with orthographic variation and sparse lexical overlap. It allows the selector to exploit subword similarity without requiring a dedicated embedding model trained for this language pair.

\paragraph{Prompt format.}
The user prompt is built as a sequence of aligned demonstrations followed by the test sentence:
\begin{quote}
\ttfamily\small
SOURCE: <src\_1>\\
RUSSIAN: <ref\_1>\\[0.25em]
$\vdots$\\[0.25em]
SOURCE: <src\_k>\\
RUSSIAN: <ref\_k>\\[0.25em]
SOURCE: <src\_test>\\
RUSSIAN:
\end{quote}
All models receive the same base system instruction:
\begin{quote}
\ttfamily\small
You are a professional translator. Translate from Komi-Yazvin to Russian. Output only the Russian translation. No explanations. No quotes. No extra lines.
\end{quote}

In the prompt itself, we retain the label ``Komi-Yazvin,'' matching the wording used in the experimental pipeline.

This prompt format was chosen to minimize ambiguity and to keep the task definition identical across providers. The demonstrations are intentionally simple parallel examples rather than more elaborate instructions, because our goal is to evaluate whether the model can infer a translation mapping from a very small aligned context. In a few cases we apply minimal model-specific wrappers only to preserve comparability, for example explicitly disabling hidden reasoning when a model family tends to consume output budget on latent deliberation instead of returning the translation.

\paragraph{Choice of models.}
We evaluate a deliberately heterogeneous set of general-purpose LLMs available through the same API interface:
\begin{itemize}
    \item \texttt{qwen/qwen3-30b-a3b}
    \item \texttt{meta-llama/llama-4-scout}
    \item \texttt{mistralai/mistral-small-3.2-24b}
    \item \texttt{google/gemini-2.5-flash-lite}
    \item \texttt{deepseek/deepseek-chat-v3.1}
    \item \texttt{qwen/qwen-2.5-72b-instruct}
    \item \texttt{anthropic/claude-sonnet-4.6}
    \item \texttt{openai/gpt-5.4}
    \item \texttt{x-ai/grok-4.20}
    \item \texttt{google/gemini-3.1-pro-preview}
\end{itemize}
The guiding principle behind model selection is not to optimize for one provider or one model family, but to cover a broad and meaningful slice of the current LLM landscape under a unified prompting and evaluation protocol. In particular, the set spans:
(i) multiple providers with different training data mixtures and alignment procedures;
(ii) both relatively cost-efficient and frontier-level models;
(iii) several model families that are widely used in multilingual or instruction-following settings; and
(iv) both open-family and proprietary systems.

This diversity matters because there is no a priori reason to expect the same architecture or provider to dominate in a severely low-resource translation task. Some models may benefit more from in-context learning, some may be stronger at strict format following, and some may have better multilingual lexical transfer. By evaluating multiple model families under the same sentence-level, story-aware protocol, we reduce the risk of drawing conclusions that are really specific to a single provider's interface or alignment behavior.

\paragraph{Evaluation metrics.}
We report corpus-level BLEU, chrF, and TER. We treat chrF as the primary metric for model ranking. The reason is practical as well as linguistic: chrF is character-based and therefore less brittle than BLEU when references and hypotheses differ in morphology, inflection, or minor orthographic detail. This is particularly important in a small test setting, where a few mismatched word forms can disproportionately affect word-level overlap metrics.

BLEU is still useful as a complementary signal because it rewards higher-precision lexical overlap and remains widely interpretable in MT research. TER provides a different perspective: rather than measuring overlap directly, it approximates the amount of editing needed to transform the hypothesis into the reference. We therefore view the three metrics as complementary. chrF is used as the primary metric because it is the most stable and appropriate for this low-resource, morphologically rich setting; BLEU and TER are reported to make the comparison more complete. Because these metrics encode different evaluative assumptions, we do not collapse them into a single scalar ranking. As an additional adequacy-oriented signal, we report LLM-as-a-judge scores produced with \texttt{openai/gpt-4.1-mini}. We also checked other GPT judge variants and obtained the same qualitative conclusions and near-identical model ordering, so we report the \texttt{gpt-4.1-mini} results throughout for consistency.

\paragraph{Why uncertainty is estimated at the story level.}
In a corpus of this size, a single aggregate score can be misleading. Moreover, sentences are not independent in the strong sense assumed by naive sentence-level resampling, because they are grouped within stories. For that reason, we estimate confidence intervals using bootstrap resampling at the story level rather than the sentence level. This yields more conservative uncertainty estimates and aligns the statistical procedure with the grouping structure that also motivated the split protocol.

\paragraph{Technical implementation and reproducibility.}
The entire experiment is implemented as a resumable pipeline rather than a one-off notebook. Each aligned sentence pair receives a stable \texttt{row\_id}. Each API request is uniquely identified by a cache key that depends on the model, fold, \texttt{row\_id}, number of demonstrations $k$, selector name, prompt version, and a hash of the concrete prompt. All requests and responses are written to append-only JSONL logs, while aggregated CSV tables and run manifests are checkpointed during execution. This design allows interrupted runs to resume safely and makes it possible to trace every final prediction back to the exact prompt and model call that produced it.

For transparency and reuse, the dataset release is hosted on \href{https://huggingface.co/datasets/pparshakov/komi-yazva-russian-parallel}{Hugging Face}, and the codebase for corpus preparation, prompting, caching, and evaluation is hosted on \href{https://github.com/pparshakov/komi-yazva}{GitHub}.

We use deterministic decoding with temperature $0.0$ and a maximum output length of 512 tokens. Hidden reasoning is disabled to prevent models from spending output budget on latent deliberation and to keep generations comparable across providers. The released implementation keeps API credentials outside the repository and records enough metadata to reproduce each prompt, response, and evaluation decision.

\paragraph{Generation validation and failure handling.}
A nontrivial part of the experimental setup concerns generation validation. In this task, some models do not always return a clean Russian translation even when prompted strictly. We therefore normalize and validate all outputs before evaluation. The validation stage detects prompt leakage (e.g., the model continues the \texttt{SOURCE/RUSSIAN} template), source copying, non-Russian outputs, transliterated or gloss-like responses, and formatting artifacts such as inline dictionaries, markdown fragments, or explanatory commentary.

This validation is essential for two reasons. First, without it, some malformed generations would be counted as translations even though they are clearly not usable outputs. Second, some apparent failures are actually technical rather than linguistic, such as transport errors or provider-side refusals. We therefore distinguish transport/API failures from content-level failures. Transport failures are retried separately, while outputs that remain malformed after validation retries are retained as errors in the final tables. This gives a cleaner picture of both translation quality and operational reliability.

\paragraph{Summary of the design rationale.}
In summary, the experimental design is built around a small but story-structured corpus, with sentence-level translation, story-level cross-validation, deterministic in-fold few-shot retrieval, multiple values of $k$, a diverse set of modern LLMs, and evaluation centered on chrF with story-level uncertainty estimates. Each component is motivated by the same objective: to obtain a comparison that is fair, reproducible, and informative under the constraints of an extremely low-resource translation setting. Section~\ref{sec:results} reports what this benchmark reveals about model quality, reliability, and sensitivity to prompting regime.

\section{Results}
\label{sec:results}

Using the evaluation protocol defined in Section~\ref{sec:experimental_design}, this section reports aggregate model performance on translation from Komi-Yazva into Russian. We begin with model-level averages, which provide a compact view of quality and reliability across the evaluated runs. In addition to reference-based MT metrics, we include an LLM-as-a-judge score as a complementary indicator of overall translation quality. Because the protocol measures multiple dimensions, we report quality, reliability, and prompt sensitivity separately rather than reducing performance to a single headline score.

\subsection{Average performance by model}

Table~\ref{tab:model-averages} summarizes mean results for each model. Higher values are better for LLM-as-a-judge, BLEU, chrF, and successful rows, whereas lower TER indicates better performance.

\begin{table}[t]
\centering
\scriptsize
\setlength{\tabcolsep}{2pt}
\begin{tabular}{L{0.34\columnwidth}rrrrr}
\toprule
Model & Judge & BLEU & chrF & TER & Success \\
\midrule
Gemini 3.1 Pro & 3.27 & 28.52 & 49.16 & 59.31 & 436.0 \\
Claude Sonnet 4.6 & 2.62 & 19.98 & 40.02 & 68.50 & 454.3 \\
Grok 4.20 & 2.43 & 15.57 & 34.48 & 74.36 & 415.7 \\
GPT-5.4 & 1.93 & 8.65 & 27.33 & 94.55 & 456.3 \\
Gemini 2.5 Flash-Lite & 1.82 & 8.45 & 23.46 & 90.37 & 457.0 \\
Llama 4 Scout & 1.75 & 6.66 & 23.47 & 93.92 & 456.7 \\
DeepSeek V3.1 & 1.75 & 4.43 & 22.37 & 155.95 & 457.0 \\
Qwen 2.5 72B & 1.66 & 6.69 & 21.89 & 94.74 & 457.0 \\
Mistral Small 3.2 24B & 1.60 & 4.87 & 20.23 & 100.46 & 457.0 \\
Qwen3 30B A3B & 1.54 & 4.31 & 18.61 & 97.65 & 457.0 \\
\bottomrule
\end{tabular}
\caption{Average results by model across the evaluated runs. ``Judge'' denotes the mean LLM-as-a-judge score. Higher is better for Judge, BLEU, chrF, and successful rows; lower is better for TER.}
\label{tab:model-averages}
\end{table}

Several patterns stand out. First, Gemini 3.1 Pro is the strongest model by all quality-oriented metrics: it achieves the highest LLM-as-a-judge score, the highest BLEU and chrF, and the lowest TER. This makes it the clearest quality leader in the comparison. At the same time, its average number of successful rows is lower than that of several competing systems, which suggests a trade-off between top-end translation quality and operational stability.

Second, Claude Sonnet 4.6 offers the strongest balance between quality and reliability. Although it does not match Gemini 3.1 Pro on BLEU or chrF, it ranks second by all main quality measures while maintaining a substantially higher average number of successful rows. In practical terms, this makes it a particularly competitive model when both translation quality and completion robustness matter.

Third, the remaining models illustrate that high completion rates do not automatically imply high translation quality. Systems such as GPT-5.4, Gemini 2.5 Flash-Lite, Llama 4 Scout, DeepSeek V3.1, Qwen 2.5 72B, Mistral Small 3.2 24B, and Qwen3 30B A3B usually return outputs for nearly all rows, but their BLEU and chrF scores remain substantially below the top-performing models. Grok 4.20 occupies an intermediate position: it is noticeably stronger than most lower-ranked models on quality, but its lower average number of successful rows indicates less stable behavior.

Finally, the ranking induced by the LLM-as-a-judge scores is broadly consistent with the reference-based metrics. This agreement is encouraging because it suggests that the automatic overlap-based measures and the judge-based evaluation capture a similar ordering of model performance in this task. Taken together, these averages indicate that zero-shot and few-shot translation quality in the Komi-Yazva setting varies sharply by model family, and that the best-performing systems are distinguished not only by higher chrF and BLEU, but also by a more favorable balance between adequacy and reliability. Put differently, the answer to ``which model is best?'' depends partly on which evaluative claim is being prioritized: raw quality, usable-output rate, or their balance.

Appendix~\ref{app:quality-vs-reliability} visualizes this quality--reliability trade-off directly. The plot makes the contrast between Gemini 3.1 Pro's quality leadership and Claude Sonnet 4.6's stronger quality--stability balance especially easy to inspect.

Appendix~\ref{app:fold-metric-summary} reports fold-level summary statistics for chrF, BLEU, TER, and success/error counts by model and prompting regime. These results complement the aggregate averages by showing how strongly each system varies across the five story-level folds.

\subsection{Average performance by number of demonstrations}

Table~\ref{tab:k-averages} summarizes the same runs aggregated by the number of in-context demonstrations rather than by model. This view is important because the experimental design explicitly compares three prompting regimes: zero-shot ($k=0$), a minimal few-shot condition ($k=4$), and a stronger few-shot condition ($k=8$). Higher values are better for LLM-as-a-judge, BLEU, chrF, and successful rows, whereas lower TER indicates better performance.

\begin{table}[t]
\centering
\scriptsize
\setlength{\tabcolsep}{4pt}
\begin{tabular}{rrrrrr}
\toprule
$k$ & Judge & BLEU & chrF & TER & Success \\
\midrule
0 & 1.54 & 6.44 & 19.72 & 97.74 & 443.3 \\
4 & 2.23 & 12.84 & 31.63 & 85.11 & 453.7 \\
8 & 2.34 & 13.15 & 32.96 & 96.09 & 454.2 \\
\bottomrule
\end{tabular}
\caption{Average results by number of demonstrations across all evaluated models and folds. ``Judge'' denotes the mean LLM-as-a-judge score. Higher is better for Judge, BLEU, chrF, and successful rows; lower is better for TER.}
\label{tab:k-averages}
\end{table}

\begin{figure}[t]
\centering
\includegraphics[width=\columnwidth]{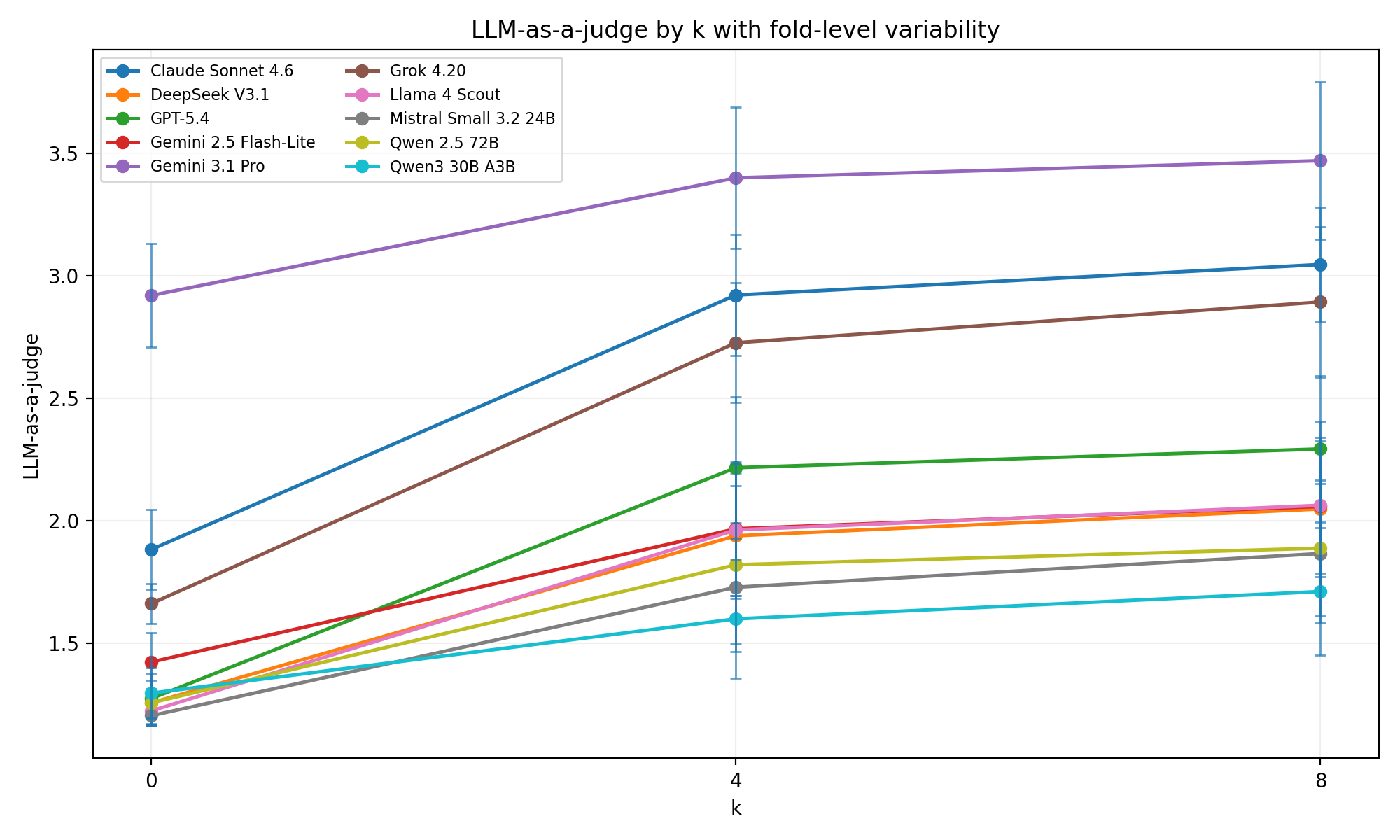}
\caption{LLM-as-a-judge score as a function of the number of in-context demonstrations for each evaluated model.}
\label{fig:judge-by-k}
\end{figure}

Figure~\ref{fig:judge-by-k} makes the same pattern visible at the model level. For nearly all systems, the largest improvement occurs when moving from zero-shot prompting to the first few retrieved demonstrations. The strongest curves belong to Gemini 3.1 Pro, Claude Sonnet 4.6, and Grok 4.20, which remain ahead of the rest across all three values of $k$. At the same time, most models show only modest additional gains between $k=4$ and $k=8$, which visually reinforces the diminishing-returns pattern already suggested by the aggregate averages.

For a complementary per-story perspective, Appendix~\ref{app:story-chrf-top-models} plots story-level chrF scores for the strongest models. This makes it easier to see where the top systems track each other closely and where their performance diverges on particular stories.

The main pattern is clear: moving from zero-shot to few-shot prompting produces a large improvement in translation quality. Relative to $k=0$, the $k=4$ condition raises the mean LLM-as-a-judge score from 1.54 to 2.23, nearly doubles BLEU from 6.44 to 12.84, and increases chrF from 19.72 to 31.63. At the same time, TER drops from 97.74 to 85.11, and the average number of successful rows rises from 443.3 to 453.7. Taken together, these shifts indicate that even a very small number of retrieved translation examples provides substantial lexical and structural guidance in this extremely low-resource setting.

The comparison between $k=4$ and $k=8$ is more nuanced. Quality continues to improve, but only modestly: the judge score rises from 2.23 to 2.34, BLEU from 12.84 to 13.15, and chrF from 31.63 to 32.96. The average number of successful rows also changes very little, increasing from 453.7 to 454.2. This suggests diminishing returns once the prompt already contains a small but informative set of demonstrations. In other words, most of the benefit of in-context learning appears to come from leaving the zero-shot regime rather than from continually increasing the number of examples.

TER introduces an important qualification. Whereas $k=4$ gives the best TER in the table, the $k=8$ condition worsens TER to 96.09 despite improving the other metrics slightly. We interpret this not as evidence that additional demonstrations are broadly harmful, but rather as a sign that the gains from larger prompts are not uniform across evaluation criteria. chrF and the judge scores suggest somewhat better adequacy and overlap at $k=8$, while TER indicates that these outputs may still require more local edits on average than those produced with $k=4$. In a small, morphologically rich corpus, such metric divergence is plausible and reinforces the decision to report multiple complementary measures rather than relying on a single score. More generally, it shows that even within one task, evaluative conclusions can shift with the metric family being emphasized.

Overall, the by-$k$ averages support the central methodological choice of the paper. Zero-shot prompting is a meaningful baseline, but it is clearly weaker than retrieval-based few-shot prompting. At the same time, the difference between $k=4$ and $k=8$ is much smaller than the difference between $k=0$ and $k=4$, which suggests that a compact few-shot context already captures most of the accessible gains. This pattern is consistent with the article's broader argument: in extremely low-resource translation, careful inference-time design matters greatly, but the marginal utility of simply adding more demonstrations appears to saturate quickly.

Appendix~\ref{app:judge-folds} provides the fold-level LLM-as-a-judge results underlying this aggregate picture. The fold breakdown shows that the jump from $k=0$ to few-shot prompting is not driven by a single favorable split: for the strongest systems, especially Gemini 3.1 Pro, Claude Sonnet 4.6, and Grok 4.20, the judge scores rise consistently across folds once retrieved demonstrations are added. At the same time, the within-fold standard deviations are typically larger in the few-shot conditions than in zero-shot mode, which suggests that demonstrations improve average quality without removing sentence-level variability.

\section{Discussion}

Taken together, the results support three substantive conclusions about LLM translation from Komi-Yazva into Russian and one methodological conclusion about evaluation in this setting. First, model choice matters more than any other single design decision in this benchmark. The gap between the strongest and weakest systems is large across judge-based and reference-based metrics alike, which suggests that multilingual coverage, instruction-following behavior, and robustness to low-resource prompting differ substantially across model families. In practical terms, the benchmark is difficult enough that strong general-purpose performance does not guarantee strong performance on this translation direction.

Second, the comparison between zero-shot and few-shot prompting shows that inference-time adaptation is genuinely useful in this setting. Retrieved demonstrations produce a substantial quality gain on average, especially when moving from $k=0$ to $k=4$. This is an important result for endangered and severely under-resourced languages, because it shows that a small curated parallel corpus can already function as a high-value prompting resource even when it is far too small for conventional model training.

Third, the gains from additional demonstrations appear to saturate quickly. The aggregate tables and the judge-by-$k$ figure both suggest that most of the accessible improvement is obtained once the model is given a small but relevant few-shot context. Increasing the number of demonstrations from $k=4$ to $k=8$ still helps some models, but the improvement is modest and not uniform across metrics. This implies that demonstration quality and retrieval relevance are likely more important than simply increasing prompt length.

More broadly, the study contributes a concrete evaluation template for extremely low-resource MT with LLMs. The combination of story-level cross-validation, deterministic in-fold retrieval, strict output validation, and model- and $k$-level aggregation provides a setup that is small in scale but methodologically disciplined. That matters because results in endangered-language translation can otherwise be dominated by accidental prompt choices, leakage across related examples, or unexamined provider-side failures. In that sense, the paper's contribution is not only a benchmark result, but an explicit, stress-testable methodology for what should count as evidence in this setting.

\section{Limitations}

The most important limitation is corpus size. Although the dataset is valuable precisely because Komi-Yazva resources are scarce, 457 sentence pairs across 74 stories still form a very small benchmark. This limits statistical power, makes individual stories relatively influential, and means that even story-level bootstrap intervals should be interpreted cautiously. The current benchmark is therefore better suited for controlled comparison than for strong claims about real-world deployment quality. More generally, the paper supports bounded evaluative claims within this protocol, not broad claims about all Komi-Yazva translation settings.

A second limitation is the absence of non-LLM translation baselines in the current study. The literature review motivates comparison against multilingual transfer and pivot-based MT systems, but the experimental section reported here focuses on prompting-based evaluation only. As a result, the paper can show relative differences across prompting regimes and model families, but it cannot yet determine whether LLM prompting is superior to stronger non-LLM baselines for this language pair.

Third, the evaluation remains partly dependent on automatic and model-based metrics. chrF, BLEU, and TER capture different aspects of overlap and edit effort, while the LLM-as-a-judge score adds a useful adequacy-oriented perspective. However, none of these measures fully replaces expert human evaluation, especially for an endangered language with limited standardization and potentially acceptable variation in lexical choice or morphology. The judge-based results should therefore be interpreted as a complementary signal rather than a definitive measure of translation quality.

A fourth limitation concerns prompt construction itself. The study uses a deterministic retrieval strategy and fixed prompt format in order to maximize reproducibility, but this also narrows the design space. Different retrieval functions, lexical grounding strategies, or richer task instructions could change the relative behavior of the models. In that sense, the reported results should be understood as strong evidence for one carefully controlled prompting setup, not as the final word on what LLM prompting can achieve for Komi-Yazva.

Finally, the operational notion of ``success'' used in the paper is intentionally strict: malformed outputs, prompt leakage, source copying, and non-Russian generations are counted as failures. This is methodologically justified, but it also means that the benchmark evaluates both translation quality and compliance with a narrowly specified output format. For practical deployment, some of these cases might be recoverable with post-editing or additional prompting stages. Future work should therefore distinguish more explicitly between recoverable formatting failures and genuine translation failures.

\section{Conclusion}

This paper makes a combined evaluation-and-dataset contribution. First, it introduces, to our knowledge, the first parallel Komi-Yazva--Russian corpus for a severely under-resourced language variety whose translation data remain extremely scarce, and it documents the scope and intended evaluative use of that resource. This corpus fills an important resource gap and provides a foundation for reproducible research on Komi-Yazva machine translation.

Second, it pairs the corpus with an explicit evaluation protocol for zero- and few-shot LLM translation, including story-level leakage control, deterministic retrieval, strict failure handling, complementary metrics, and story-level uncertainty estimation. Under this protocol, modern large language models can produce non-trivial translations from Komi-Yazva into Russian, but the conclusions are necessarily qualified. Retrieval-based few-shot prompting is consistently stronger than zero-shot prompting, while the gains from simply increasing the number of examples remain limited. At the same time, model rankings depend partly on whether one emphasizes raw quality, reliability, or edit-based effort, which reinforces the paper's broader point that evaluation design shapes scientific claims. Overall, the work offers not a new model, but a reproducible dataset-and-evaluation setup for studying endangered-language translation more carefully.

\bibliography{custom}

\clearpage
\onecolumn
\appendix

\section{Representative low-resource LLM translation studies}
\label{app:lrl-matrix}

Table~\ref{tab:lrl-llm-matrix} summarizes representative empirical studies aligned with underrepresented-language translation and the few-hundred to sub-10k parallel-sentence regime.

\begin{table}[!t]
\raggedright
\small
\setlength{\tabcolsep}{3.5pt}
\begin{tabular}{L{0.17\linewidth}L{0.18\linewidth}L{0.18\linewidth}L{0.37\linewidth}}
\hline
\textbf{Study} & \textbf{Resource regime} & \textbf{Prompting regime} & \textbf{Key empirical finding} \\
\hline
\citet{tereshchenko-etal-2025-gpt-uralic} & Endangered Uralic directions including Finnish--Komi-Zyrian & Prompted GPT translation with reasoning vs non-reasoning architectures & Reasoning models reduce refusal and improve practical usability for endangered-language translation workflows. \\
\citet{rueter-etal-2020-komi-permyak} and \citet{partanen-etal-2018-ud-komi} & Komi-Permyak/Komi-Zyrian infrastructure and treebanks (resource-building regime) & Not a prompting study & Resource and annotation infrastructure is mature for some Komi varieties, but direct Komi-Yazva MT evaluation remains underdeveloped. \\
\citet{thakur-2024-no-resource} & No-resource setting (often $<100$ pairs), Owens Valley Paiute case study & Direct and reasoning-based prompting in {LLM}s & In no-resource conditions, in-context prompting is often more viable than standard low-resource fine-tuning. \\
\citet{merx-etal-2024-mambai} & English--Mambai, few-shot and dictionary-supported & Retrieval-augmented few-shot prompting with lexical entries & Retrieval plus dictionary cues improve translation quality and stability across test sets. \\
\citet{elsner-needle-2023-translating} & English--Inuktitut (unwritten), effectively no parallel bitext & {GPT}-3 prompting with dictionary and grammar cues & Dictionary-grounded prompting yields usable output without bitext, but robustness remains limited. \\
\citet{zhang-etal-2024-unseen} & Unseen-language adaptation with only dictionary support and around 5k sentence pairs in reported settings & On-the-fly in-context adaptation framework for unseen languages & Structured on-the-fly adaptation substantially improves translation for languages absent from pretraining coverage. \\
\citet{pei-etal-2025-understanding} & Manchu with 3{,}520 real parallel pairs (+ monolingual augmentation) & In-context prompting with dictionaries and retrieved demonstrations & Dictionary and nearest-example cues are highly effective; prompting-generated data improves low-resource fine-tuning. \\
\citet{zebaze-etal-2025-context} & English to low-resource targets (e.g., Hausa/Javanese/Wolof) & 1/5/10-shot prompting with semantic retrieval & Similarity-based retrieval improves quality and robustness under noisy candidate pools. \\
\citet{zebaze-etal-2025-compositional} & One-side-low-resource pairs with small retrieval pools (e.g., 997 and 971 examples) & Compositional prompting with decomposition and retrieved evidence & Compositional prompting improves quality and out-of-domain robustness. \\
\citet{hendy-etal-2023-gpt} & 18 directions including low-resource and non-English-centric settings & Zero-shot and few-shot prompting across {GPT} families & {GPT} models are competitive on high-resource pairs but degrade in low-resource and non-English-centric directions. \\
\citet{ramasethu-etal-2026-pivot} & Extremely low-resource setup with linguistically related pivots & Few-shot prompting with related-language pivot guidance & Pivot prompting can help in some language pairs but gains are modest and configuration-sensitive. \\
\citet{alves-etal-2023-steering} & Ten language pairs with varied resource levels & Few-shot in-context prompting with adaptation & Fine-tuning usually outperforms prompting alone; hybrid strategies keep controllability. \\
\citet{zhang-etal-2023-machine} & {WMT} setup (mostly high-resource French--English) as methodological reference & Zero-shot and few-shot prompting with {LLM}s & {QL}o{RA} fine-tuning strongly outperforms prompting-only strategies. \\
\citet{ahmed-buys-2024-neural} & Very low-resource Southern African language pairs & Not prompt-based (transfer/pivot baseline) & Synthetic pivoting yields strong gains (up to +5.6 {BLEU}) in very low-resource settings. \\
\hline
\end{tabular}
\caption{Empirical studies on prompt-based {LLM} translation for low-resource settings with fine-tuning and transfer/pivot comparisons.}
\label{tab:lrl-llm-matrix}
\end{table}

\section{Story-level chrF for top models}
\label{app:story-chrf-top-models}

\begin{figure}[!t]
\centering
\includegraphics[width=0.92\textwidth]{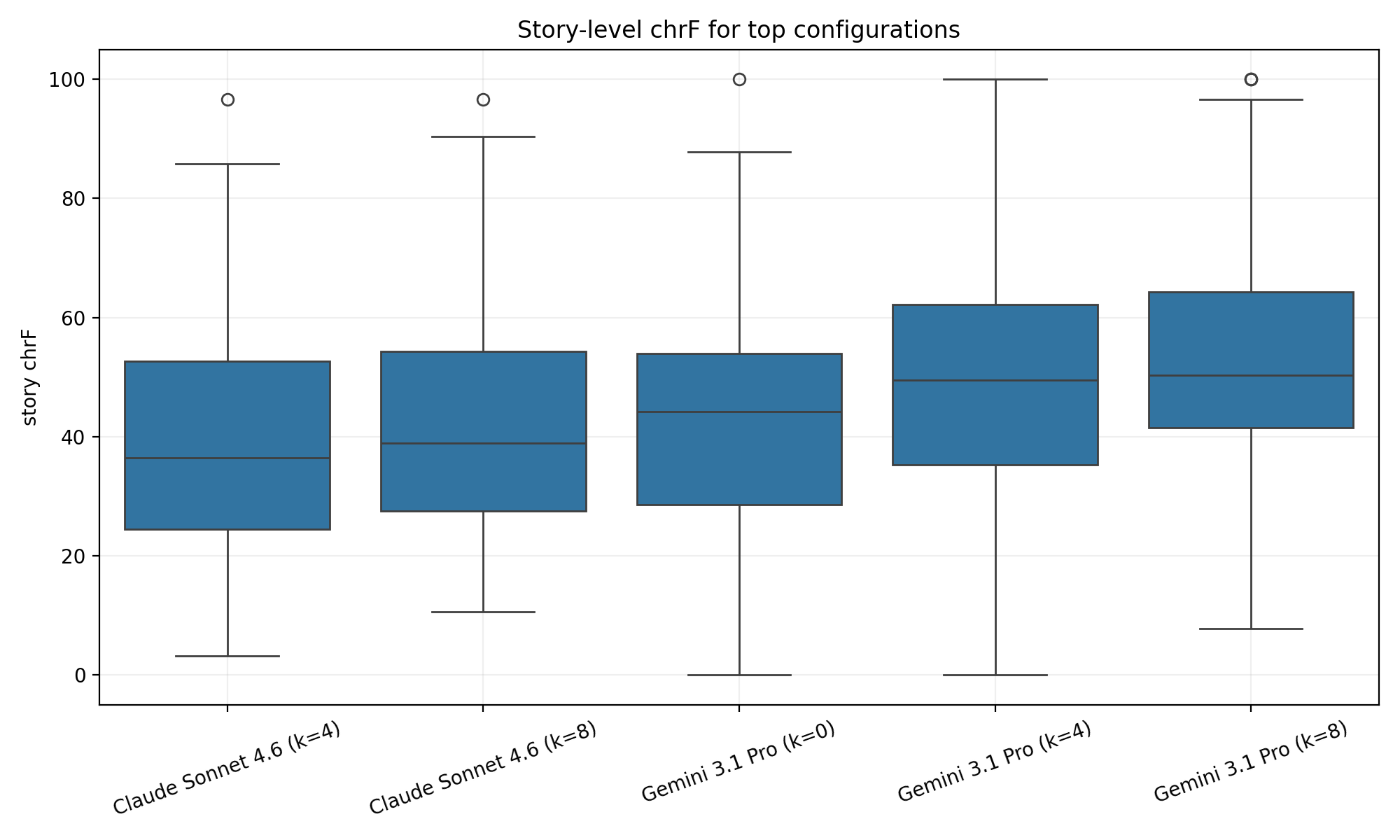}
\caption{Story-level chrF scores for the top-performing models.}
\label{fig:story-chrf-top-models}
\end{figure}

Figure~\ref{fig:story-chrf-top-models} shows story-level chrF scores for the strongest models in the benchmark. It complements the aggregate tables by revealing how relative model ranking behaves across individual stories rather than only in the overall average.

\clearpage

\section{Quality--reliability trade-off across models}
\label{app:quality-vs-reliability}

\begin{figure}[!t]
\centering
\includegraphics[width=0.92\textwidth]{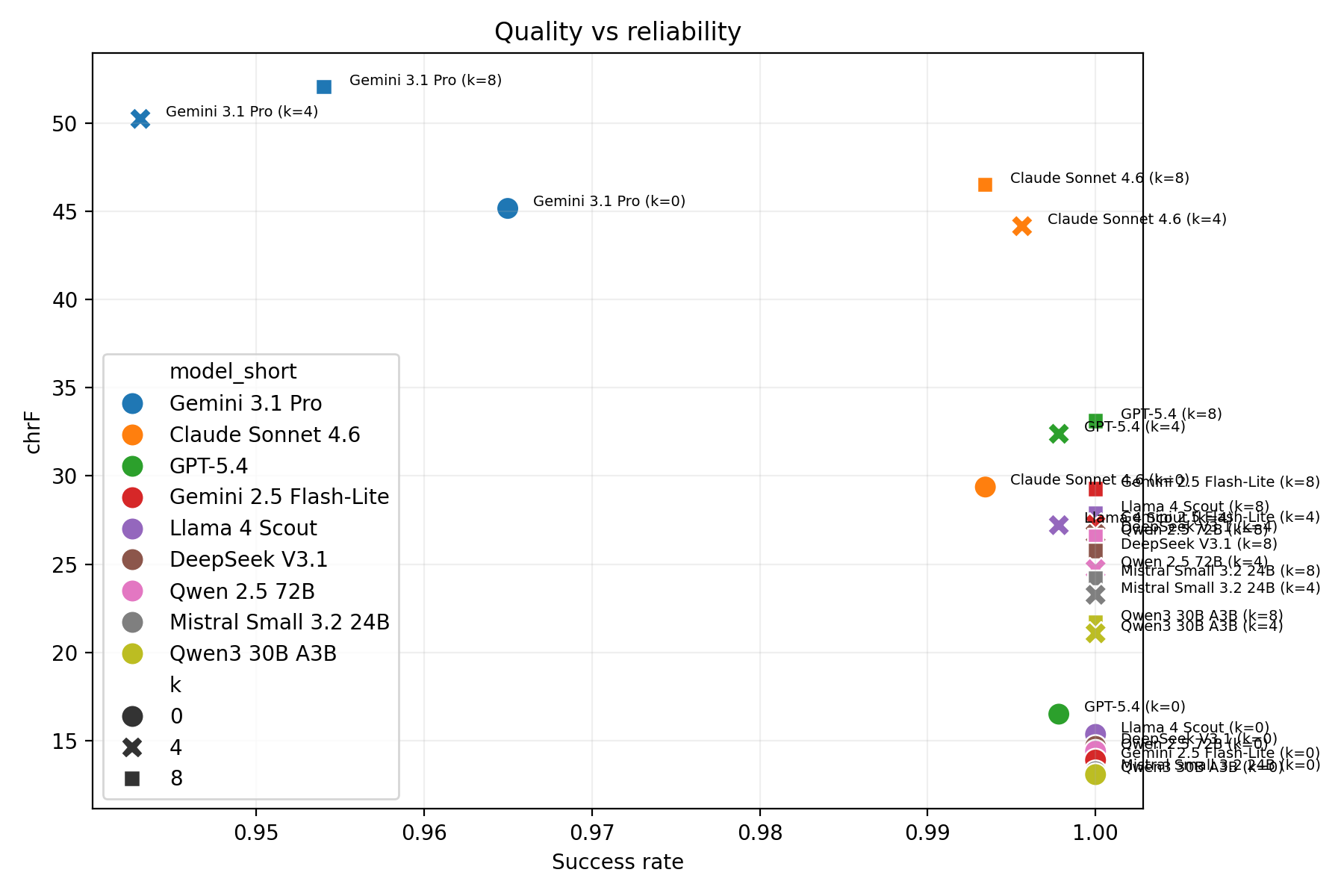}
\caption{Average translation quality versus reliability across models.}
\label{fig:quality-vs-reliability}
\end{figure}

Figure~\ref{fig:quality-vs-reliability} plots average translation quality against average reliability for all evaluated models. This appendix makes the practical trade-off discussed in the main text easier to interpret at a glance.

\clearpage

\section{Fold-level metric summary by model and prompting regime}
\label{app:fold-metric-summary}

Table~\ref{tab:fold-metric-summary} reports fold-level summary statistics for chrF, BLEU, TER, and the number of successful and error rows for each model and each value of $k$. For compactness, the table uses the short model names reported in the main text.

\tiny
\setlength{\LTleft}{0pt}
\setlength{\LTright}{0pt}
\begin{longtable}{L{0.21\textwidth}r rrrrrrrr}
\caption{Fold-level summary statistics by model and prompting regime.}\label{tab:fold-metric-summary}\\
\toprule
Model & $k$ & chrF & SD & Min & Max & BLEU & TER & Succ. & Err. \\
\midrule
\endfirsthead
\toprule
Model & $k$ & chrF & SD & Min & Max & BLEU & TER & Succ. & Err. \\
\midrule
\endhead
\midrule
\multicolumn{10}{r}{Continued on next page} \\
\midrule
\endfoot
\bottomrule
\endlastfoot
Gemini 3.1 Pro & 8 & 52.069 & 3.759 & 47.847 & 58.179 & 29.230 & 56.405 & 87.200 & 4.200 \\
Gemini 3.1 Pro & 4 & 50.328 & 4.316 & 44.730 & 55.383 & 29.011 & 56.877 & 86.200 & 5.200 \\
Claude Sonnet 4.6 & 8 & 46.441 & 6.200 & 39.876 & 53.173 & 24.650 & 59.605 & 90.800 & 0.600 \\
Gemini 3.1 Pro & 0 & 45.020 & 3.456 & 41.897 & 48.840 & 25.119 & 65.135 & 88.200 & 3.200 \\
Claude Sonnet 4.6 & 4 & 44.116 & 4.614 & 38.983 & 48.602 & 22.062 & 62.633 & 91.000 & 0.400 \\
Grok 4.20 & 8 & 42.528 & 5.204 & 37.387 & 49.071 & 20.105 & 64.509 & 90.600 & 0.800 \\
Grok 4.20 & 4 & 39.162 & 4.208 & 34.517 & 44.021 & 17.728 & 68.352 & 90.800 & 0.600 \\
GPT-5.4 & 8 & 33.248 & 4.966 & 27.188 & 40.066 & 10.472 & 103.904 & 91.400 & 0.000 \\
GPT-5.4 & 4 & 32.480 & 4.522 & 27.089 & 39.300 & 12.230 & 83.496 & 91.200 & 0.200 \\
Gemini 2.5 Flash-Lite & 8 & 29.223 & 5.182 & 23.029 & 36.635 & 10.809 & 77.423 & 91.400 & 0.000 \\
Claude Sonnet 4.6 & 0 & 29.171 & 4.083 & 24.370 & 34.184 & 11.439 & 83.816 & 90.800 & 0.600 \\
Llama 4 Scout & 8 & 28.076 & 4.406 & 22.807 & 33.812 & 8.702 & 92.786 & 91.400 & 0.000 \\
Llama 4 Scout & 4 & 27.340 & 4.120 & 22.337 & 32.404 & 8.642 & 87.574 & 91.200 & 0.200 \\
Gemini 2.5 Flash-Lite & 4 & 27.283 & 3.938 & 22.391 & 32.894 & 10.190 & 80.253 & 91.400 & 0.000 \\
DeepSeek V3.1 & 4 & 26.806 & 4.516 & 20.810 & 33.382 & 6.258 & 136.796 & 91.400 & 0.000 \\
Qwen 2.5 72B & 8 & 26.656 & 4.833 & 21.800 & 33.476 & 9.250 & 83.847 & 91.400 & 0.000 \\
DeepSeek V3.1 & 8 & 25.940 & 5.110 & 18.575 & 31.833 & 4.653 & 220.571 & 91.400 & 0.000 \\
Qwen 2.5 72B & 4 & 24.732 & 5.117 & 19.679 & 32.758 & 7.744 & 84.757 & 91.400 & 0.000 \\
Mistral Small 3.2 24B & 8 & 24.208 & 3.521 & 20.420 & 29.134 & 6.551 & 103.250 & 91.400 & 0.000 \\
Mistral Small 3.2 24B & 4 & 23.330 & 3.140 & 19.468 & 27.342 & 5.534 & 98.523 & 91.400 & 0.000 \\
Qwen3 30B A3B & 8 & 21.930 & 4.144 & 17.282 & 27.797 & 5.306 & 101.369 & 91.400 & 0.000 \\
Grok 4.20 & 0 & 21.361 & 4.766 & 14.616 & 26.404 & 6.955 & 90.853 & 68.000 & 23.400 \\
Qwen3 30B A3B & 4 & 21.208 & 3.468 & 18.171 & 26.623 & 5.565 & 92.732 & 91.400 & 0.000 \\
GPT-5.4 & 0 & 16.416 & 2.554 & 12.397 & 18.734 & 3.478 & 97.571 & 91.200 & 0.200 \\
Llama 4 Scout & 0 & 15.272 & 1.966 & 11.971 & 16.674 & 2.641 & 99.649 & 91.400 & 0.000 \\
DeepSeek V3.1 & 0 & 14.641 & 1.647 & 13.020 & 16.839 & 2.543 & 116.076 & 91.400 & 0.000 \\
Qwen 2.5 72B & 0 & 14.333 & 2.372 & 10.399 & 16.182 & 2.289 & 114.904 & 91.400 & 0.000 \\
Gemini 2.5 Flash-Lite & 0 & 13.846 & 1.695 & 11.350 & 16.022 & 3.083 & 112.231 & 91.400 & 0.000 \\
Mistral Small 3.2 24B & 0 & 13.166 & 1.637 & 10.592 & 14.397 & 2.221 & 98.895 & 91.400 & 0.000 \\
Qwen3 30B A3B & 0 & 13.066 & 1.564 & 11.042 & 15.204 & 2.394 & 97.866 & 91.400 & 0.000 \\
\end{longtable}
\normalsize

\section{Fold-level LLM-as-a-judge results}
\label{app:judge-folds}

Table~\ref{tab:judge-fold-breakdown} reports fold-level LLM-as-a-judge means, within-fold standard deviations, and the number of judged examples for every model and prompting regime. This appendix is included to make the stability patterns behind the aggregate tables fully transparent.

\tiny
\setlength{\LTleft}{0pt}
\setlength{\LTright}{0pt}
\begin{longtable}{L{0.34\textwidth}rrrrr}
\caption{Fold-level judge results by model, $k$, and fold.}\label{tab:judge-fold-breakdown}\\
\toprule
Model & $k$ & fold & Judge & Judge SD & $n$ \\
\midrule
\endfirsthead
\toprule
Model & $k$ & fold & Judge & Judge SD & $n$ \\
\midrule
\endhead
\midrule
\multicolumn{6}{r}{Continued on next page} \\
\midrule
\endfoot
\bottomrule
\endlastfoot
Claude Sonnet 4.6 & 0 & 0 & 1.73 & 0.85 & 92 \\
Claude Sonnet 4.6 & 0 & 1 & 1.98 & 1.27 & 92 \\
Claude Sonnet 4.6 & 0 & 2 & 1.78 & 1.07 & 91 \\
Claude Sonnet 4.6 & 0 & 3 & 2.12 & 1.23 & 91 \\
Claude Sonnet 4.6 & 0 & 4 & 1.81 & 1.15 & 91 \\
Claude Sonnet 4.6 & 4 & 0 & 2.95 & 1.21 & 92 \\
Claude Sonnet 4.6 & 4 & 1 & 3.20 & 1.46 & 92 \\
Claude Sonnet 4.6 & 4 & 2 & 2.68 & 1.29 & 91 \\
Claude Sonnet 4.6 & 4 & 3 & 3.13 & 1.23 & 91 \\
Claude Sonnet 4.6 & 4 & 4 & 2.66 & 1.34 & 91 \\
Claude Sonnet 4.6 & 8 & 0 & 3.09 & 1.25 & 92 \\
Claude Sonnet 4.6 & 8 & 1 & 3.33 & 1.45 & 92 \\
Claude Sonnet 4.6 & 8 & 2 & 2.88 & 1.32 & 91 \\
Claude Sonnet 4.6 & 8 & 3 & 3.20 & 1.28 & 91 \\
Claude Sonnet 4.6 & 8 & 4 & 2.75 & 1.37 & 91 \\
DeepSeek V3.1 & 0 & 0 & 1.20 & 0.54 & 92 \\
DeepSeek V3.1 & 0 & 1 & 1.36 & 0.87 & 92 \\
DeepSeek V3.1 & 0 & 2 & 1.16 & 0.60 & 91 \\
DeepSeek V3.1 & 0 & 3 & 1.22 & 0.57 & 91 \\
DeepSeek V3.1 & 0 & 4 & 1.35 & 0.81 & 91 \\
DeepSeek V3.1 & 4 & 0 & 1.98 & 1.10 & 92 \\
DeepSeek V3.1 & 4 & 1 & 2.34 & 1.51 & 92 \\
DeepSeek V3.1 & 4 & 2 & 1.92 & 1.22 & 91 \\
DeepSeek V3.1 & 4 & 3 & 1.81 & 1.07 & 91 \\
DeepSeek V3.1 & 4 & 4 & 1.65 & 1.10 & 91 \\
DeepSeek V3.1 & 8 & 0 & 2.09 & 1.15 & 92 \\
DeepSeek V3.1 & 8 & 1 & 2.38 & 1.38 & 92 \\
DeepSeek V3.1 & 8 & 2 & 1.93 & 1.15 & 91 \\
DeepSeek V3.1 & 8 & 3 & 2.20 & 1.24 & 91 \\
DeepSeek V3.1 & 8 & 4 & 1.65 & 1.07 & 91 \\
GPT-5.4 & 0 & 0 & 1.17 & 0.48 & 92 \\
GPT-5.4 & 0 & 1 & 1.37 & 0.87 & 92 \\
GPT-5.4 & 0 & 2 & 1.16 & 0.54 & 91 \\
GPT-5.4 & 0 & 3 & 1.30 & 0.61 & 91 \\
GPT-5.4 & 0 & 4 & 1.37 & 0.81 & 91 \\
GPT-5.4 & 4 & 0 & 2.28 & 1.30 & 92 \\
GPT-5.4 & 4 & 1 & 2.62 & 1.50 & 92 \\
GPT-5.4 & 4 & 2 & 2.20 & 1.34 & 91 \\
GPT-5.4 & 4 & 3 & 2.18 & 1.23 & 91 \\
GPT-5.4 & 4 & 4 & 1.81 & 1.05 & 91 \\
GPT-5.4 & 8 & 0 & 2.38 & 1.37 & 92 \\
GPT-5.4 & 8 & 1 & 2.72 & 1.47 & 92 \\
GPT-5.4 & 8 & 2 & 2.16 & 1.25 & 91 \\
GPT-5.4 & 8 & 3 & 2.31 & 1.35 & 91 \\
GPT-5.4 & 8 & 4 & 1.90 & 1.19 & 91 \\
Gemini 2.5 Flash-Lite & 0 & 0 & 1.35 & 0.93 & 92 \\
Gemini 2.5 Flash-Lite & 0 & 1 & 1.40 & 1.03 & 92 \\
Gemini 2.5 Flash-Lite & 0 & 2 & 1.63 & 1.36 & 91 \\
Gemini 2.5 Flash-Lite & 0 & 3 & 1.33 & 0.76 & 91 \\
Gemini 2.5 Flash-Lite & 0 & 4 & 1.42 & 0.97 & 91 \\
Gemini 2.5 Flash-Lite & 4 & 0 & 1.91 & 1.09 & 92 \\
Gemini 2.5 Flash-Lite & 4 & 1 & 2.38 & 1.47 & 92 \\
Gemini 2.5 Flash-Lite & 4 & 2 & 1.97 & 1.24 & 91 \\
Gemini 2.5 Flash-Lite & 4 & 3 & 1.97 & 1.11 & 91 \\
Gemini 2.5 Flash-Lite & 4 & 4 & 1.62 & 1.00 & 91 \\
Gemini 2.5 Flash-Lite & 8 & 0 & 2.25 & 1.22 & 92 \\
Gemini 2.5 Flash-Lite & 8 & 1 & 2.50 & 1.46 & 92 \\
Gemini 2.5 Flash-Lite & 8 & 2 & 1.95 & 1.28 & 91 \\
Gemini 2.5 Flash-Lite & 8 & 3 & 2.03 & 1.13 & 91 \\
Gemini 2.5 Flash-Lite & 8 & 4 & 1.57 & 0.93 & 91 \\
Gemini 3.1 Pro & 0 & 0 & 2.87 & 1.38 & 92 \\
Gemini 3.1 Pro & 0 & 1 & 3.09 & 1.38 & 92 \\
Gemini 3.1 Pro & 0 & 2 & 2.93 & 1.42 & 91 \\
Gemini 3.1 Pro & 0 & 3 & 3.12 & 1.36 & 91 \\
Gemini 3.1 Pro & 0 & 4 & 2.59 & 1.35 & 91 \\
Gemini 3.1 Pro & 4 & 0 & 3.70 & 1.17 & 92 \\
Gemini 3.1 Pro & 4 & 1 & 3.64 & 1.39 & 92 \\
Gemini 3.1 Pro & 4 & 2 & 3.26 & 1.32 & 91 \\
Gemini 3.1 Pro & 4 & 3 & 3.42 & 1.26 & 91 \\
Gemini 3.1 Pro & 4 & 4 & 2.99 & 1.39 & 91 \\
Gemini 3.1 Pro & 8 & 0 & 3.48 & 1.37 & 92 \\
Gemini 3.1 Pro & 8 & 1 & 3.97 & 1.04 & 92 \\
Gemini 3.1 Pro & 8 & 2 & 3.43 & 1.30 & 91 \\
Gemini 3.1 Pro & 8 & 3 & 3.42 & 1.32 & 91 \\
Gemini 3.1 Pro & 8 & 4 & 3.07 & 1.25 & 91 \\
Grok 4.20 & 0 & 0 & 1.64 & 1.17 & 92 \\
Grok 4.20 & 0 & 1 & 1.61 & 0.99 & 92 \\
Grok 4.20 & 0 & 2 & 1.57 & 1.02 & 91 \\
Grok 4.20 & 0 & 3 & 1.77 & 1.15 & 91 \\
Grok 4.20 & 0 & 4 & 1.73 & 1.08 & 91 \\
Grok 4.20 & 4 & 0 & 2.86 & 1.37 & 92 \\
Grok 4.20 & 4 & 1 & 3.01 & 1.44 & 92 \\
Grok 4.20 & 4 & 2 & 2.62 & 1.33 & 91 \\
Grok 4.20 & 4 & 3 & 2.78 & 1.24 & 91 \\
Grok 4.20 & 4 & 4 & 2.37 & 1.34 & 91 \\
Grok 4.20 & 8 & 0 & 2.92 & 1.26 & 92 \\
Grok 4.20 & 8 & 1 & 3.21 & 1.43 & 92 \\
Grok 4.20 & 8 & 2 & 2.79 & 1.34 & 91 \\
Grok 4.20 & 8 & 3 & 3.12 & 1.26 & 91 \\
Grok 4.20 & 8 & 4 & 2.43 & 1.31 & 91 \\
Llama 4 Scout & 0 & 0 & 1.21 & 0.64 & 92 \\
Llama 4 Scout & 0 & 1 & 1.20 & 0.56 & 92 \\
Llama 4 Scout & 0 & 2 & 1.19 & 0.61 & 91 \\
Llama 4 Scout & 0 & 3 & 1.21 & 0.51 & 91 \\
Llama 4 Scout & 0 & 4 & 1.33 & 0.84 & 91 \\
Llama 4 Scout & 4 & 0 & 2.07 & 1.14 & 92 \\
Llama 4 Scout & 4 & 1 & 2.36 & 1.47 & 92 \\
Llama 4 Scout & 4 & 2 & 1.90 & 1.12 & 91 \\
Llama 4 Scout & 4 & 3 & 1.86 & 1.16 & 91 \\
Llama 4 Scout & 4 & 4 & 1.64 & 0.96 & 91 \\
Llama 4 Scout & 8 & 0 & 2.26 & 1.36 & 92 \\
Llama 4 Scout & 8 & 1 & 2.42 & 1.38 & 92 \\
Llama 4 Scout & 8 & 2 & 1.95 & 1.09 & 91 \\
Llama 4 Scout & 8 & 3 & 1.97 & 1.22 & 91 \\
Llama 4 Scout & 8 & 4 & 1.73 & 1.02 & 91 \\
Mistral Small 3.2 24B & 0 & 0 & 1.17 & 0.69 & 92 \\
Mistral Small 3.2 24B & 0 & 1 & 1.27 & 0.71 & 92 \\
Mistral Small 3.2 24B & 0 & 2 & 1.16 & 0.75 & 91 \\
Mistral Small 3.2 24B & 0 & 3 & 1.22 & 0.63 & 91 \\
Mistral Small 3.2 24B & 0 & 4 & 1.20 & 0.64 & 91 \\
Mistral Small 3.2 24B & 4 & 0 & 1.79 & 1.05 & 92 \\
Mistral Small 3.2 24B & 4 & 1 & 2.13 & 1.40 & 92 \\
Mistral Small 3.2 24B & 4 & 2 & 1.70 & 1.08 & 91 \\
Mistral Small 3.2 24B & 4 & 3 & 1.60 & 0.85 & 91 \\
Mistral Small 3.2 24B & 4 & 4 & 1.42 & 0.76 & 91 \\
Mistral Small 3.2 24B & 8 & 0 & 1.98 & 1.15 & 92 \\
Mistral Small 3.2 24B & 8 & 1 & 2.27 & 1.43 & 92 \\
Mistral Small 3.2 24B & 8 & 2 & 1.75 & 1.02 & 91 \\
Mistral Small 3.2 24B & 8 & 3 & 1.84 & 1.08 & 91 \\
Mistral Small 3.2 24B & 8 & 4 & 1.51 & 0.86 & 91 \\
Qwen 2.5 72B & 0 & 0 & 1.26 & 0.85 & 92 \\
Qwen 2.5 72B & 0 & 1 & 1.25 & 0.72 & 92 \\
Qwen 2.5 72B & 0 & 2 & 1.18 & 0.53 & 91 \\
Qwen 2.5 72B & 0 & 3 & 1.26 & 0.65 & 91 \\
Qwen 2.5 72B & 0 & 4 & 1.34 & 0.76 & 91 \\
Qwen 2.5 72B & 4 & 0 & 1.85 & 1.16 & 92 \\
Qwen 2.5 72B & 4 & 1 & 2.35 & 1.48 & 92 \\
Qwen 2.5 72B & 4 & 2 & 1.57 & 0.93 & 91 \\
Qwen 2.5 72B & 4 & 3 & 1.79 & 1.02 & 91 \\
Qwen 2.5 72B & 4 & 4 & 1.55 & 0.97 & 91 \\
Qwen 2.5 72B & 8 & 0 & 1.99 & 1.11 & 92 \\
Qwen 2.5 72B & 8 & 1 & 2.33 & 1.44 & 92 \\
Qwen 2.5 72B & 8 & 2 & 1.75 & 1.02 & 91 \\
Qwen 2.5 72B & 8 & 3 & 1.76 & 0.94 & 91 \\
Qwen 2.5 72B & 8 & 4 & 1.63 & 1.08 & 91 \\
Qwen3 30B A3B & 0 & 0 & 1.22 & 0.77 & 92 \\
Qwen3 30B A3B & 0 & 1 & 1.33 & 0.84 & 92 \\
Qwen3 30B A3B & 0 & 2 & 1.33 & 0.97 & 91 \\
Qwen3 30B A3B & 0 & 3 & 1.18 & 0.74 & 91 \\
Qwen3 30B A3B & 0 & 4 & 1.44 & 1.04 & 91 \\
Qwen3 30B A3B & 4 & 0 & 1.68 & 1.08 & 92 \\
Qwen3 30B A3B & 4 & 1 & 1.99 & 1.37 & 92 \\
Qwen3 30B A3B & 4 & 2 & 1.49 & 0.89 & 91 \\
Qwen3 30B A3B & 4 & 3 & 1.44 & 0.65 & 91 \\
Qwen3 30B A3B & 4 & 4 & 1.40 & 0.85 & 91 \\
Qwen3 30B A3B & 8 & 0 & 1.79 & 1.10 & 92 \\
Qwen3 30B A3B & 8 & 1 & 2.10 & 1.33 & 92 \\
Qwen3 30B A3B & 8 & 2 & 1.70 & 1.05 & 91 \\
Qwen3 30B A3B & 8 & 3 & 1.56 & 0.86 & 91 \\
Qwen3 30B A3B & 8 & 4 & 1.41 & 0.91 & 91 \\
\end{longtable}
\normalsize

\end{document}